\documentclass[10pt,journal,compsoc]{IEEEtran}
\ifCLASSOPTIONcompsoc
 \usepackage[nocompress]{cite}
\else
 \usepackage{cite}
\fi

\ifCLASSINFOpdf
\else
\fi
\usepackage{algorithmic}
\ifCLASSOPTIONcompsoc
 \usepackage[font=normalsize,labelfont=sf,textfont=sf]{subfig}
\else
 \usepackage[font=footnotesize]{subfig}
\fi
\ifCLASSOPTIONcompsoc
 \usepackage[font=normalsize,labelfont=sf,textfont=sf]{subfig}
\else
 \usepackage[font=footnotesize]{subfig}
\fi

\usepackage{amsmath,graphicx,multirow}
\usepackage{bbm,amssymb}
\usepackage{color}

\usepackage[ruled,lined,linesnumbered,boxed]{algorithm2e}
\usepackage{algorithmic}

\def\K{{\bf K}}

\def\P{{\bf P}}
\def\t{{\bf \textrm{tr}}}
\def\Y{{\bf Y}} 
\def\x{{\bf x}} 
\def\S{{\cal S}}

\def\KK{{\cal K}}

\def\ds1{\mathds{1}}
\def\PPhi{{\bf \Phi}}

\ifCLASSINFOpdf
\else
\fi
\begin{document}
\title{Learning Deep Context-Network Architectures for Image Annotation}

\author{Mingyuan Jiu and Hichem Sahbi
  \IEEEcompsocitemizethanks{\IEEEcompsocthanksitem Mingyuan Jiu is with the School of Information Engineering, Zhengzhou University, Zhengzhou, 450001, China. Email: iemyjiu@zzu.edu.cn.  Mingyuan Jiu was with T\'el\'ecom ParisTech as a postdoc, working with Hichem Sahbi, when this work was discussed and part of it achieved and written (under the MLVIS project).\protect    
\IEEEcompsocthanksitem Hichem Sahbi is with the CNRS, UPMC Sorbonne University, Paris, France. E-mail: hichem.sahbi@lip6.fr \protect}
\thanks{}}

\IEEEcompsoctitleabstractindextext{
\begin{abstract}
Context plays an important role in   visual pattern recognition as it provides complementary clues for  different learning   tasks including image classification and  annotation.  In the particular scenario of kernel learning, the general recipe  of context-based kernel design consists in learning positive semi-definite similarity functions that return high values not only when data share similar content but also similar context. However, in spite of having a positive impact on performance, the use of context in these kernel design methods has not been fully explored; indeed, context has been handcrafted instead of being learned.\\
In this paper, we introduce a novel context-aware kernel design framework based on deep learning. Our method discriminatively learns spatial geometric context as the weights of a deep network (DN). The architecture of this network is fully determined by the solution of an objective function that mixes content, context and regularization, while the parameters of this network determine the most relevant (discriminant) parts of the learned context.  We apply this context and kernel learning framework to image classification using the challenging ImageCLEF Photo Annotation benchmark; the latter shows that our deep context learning    provides highly effective kernels for image classification as corroborated through extensive experiments.  
\end{abstract}

}
\maketitle

\IEEEdisplaynontitleabstractindextext
\IEEEpeerreviewmaketitle

\section{Introduction} \label{sec:intro}

Image annotation is a major challenge in computer vision, which aims to describe  pictures with  multiple   semantic concepts taken from  large vocabularies (also known as keywords, classes or categories)  \cite{Bernard2003, Makadia2008,boujemaa2004visual,vo2012transductive,sahbi2013cnrs}. This problem is  motivated by different visual recognition  tasks (including multimedia information access, human computer interaction, robotics, autonomous driving, etc.) and also by the exponential growth of visual contents in the web and  social medias. Image annotation is also very challenging due to the difficulty to characterize and discriminate a large number of highly variable concepts (including abstract notions) which are widely used in visual recognition.\\  
\indent Existing annotation methods are usually based on machine learning; first, they learn intricate relationships between concepts and image features (e.g.~color, shape,  deep features, etc.) using different classifiers (SVMs, \cite{Goh2005, Qi2007,sahbi2002coarse}, nearest neighbors~\cite{Guillaumin2009, Verma2012}, etc.), then they assign concepts to new images depending on the scores of the learned classifiers.   Among these classification techniques, SVMs are known to be effective in image annotation \cite{Villegas2013}, but their success is highly dependent on the choice of kernels \cite{ShaweTaylor2004}. The latter, defined as symmetric and positive semi-definite functions, should provide high values when images share similar semantic contents and vice-versa; among existing kernels, linear, polynomial, gaussian and histogram intersection are particularly popular.  In addition to these widely used standard kernels, many existing algorithms have been proposed in the literature in order to combine and learn  discriminant kernels (from these standard functions) that capture better semantic similarity such as multiple kernel learning \cite{Bach2004, Jiu2015, Jiu2016a} and additive kernels \cite{Vedaldi2012}. However, all these standard kernels and their combinations rely   on the local visual content of images, which is clearly insufficient  to capture their semantic  especially when this content is highly variable.  In this work, we focus on learning better kernels with a high discrimination power, by including {\it not only content but also context} which is  also learned instead of being handcrafted. Moreover, the design principle of our method allows us to obtain the explicit form of the learned kernels and this makes their evaluation highly efficient, especially for large scale databases. 

Considering a collection of images, each one seen as a constellation of cells (grid of patches \cite{thiemert2005applying}) with  each cell encoded with appearance features (e.g., Bag-of-Words, deep VGG-Net features, etc.), our goal is to learn  accurate  kernel functions that measure similarity between images by comparing their constellations of cells. The design principle  of our method  consists in learning appropriate kernels (for cell comparison)  that return relevant values;  this is achieved not only by comparing content of cells, but also their spatial geometric context which provides  complementary clues for classification (see also \cite{Belongie01shapematching, Hecvpr2004, Sahbi2011, Jiuprl2014,li2011superpixel}). In our proposed solution (in section \ref{sec:unslearning}), two cells\footnote{belonging to two images.} with similar visual contents should be declared as different if their contexts (i.e., their spatial neighborhoods) are different while two cells with different visual contents could be declared as similar if their contexts are  similar. In this paper, we implement this principle by designing context-aware kernels as the solution of  an optimization  problem which includes three criteria: a fidelity term  that measures the similarity between cells using local content only, a context term that strengthens (or weakens) the similarity between cells depending on their spatial neighbors and finally a regularization criterion that controls the smoothness  of the learned kernel values.  The initial formulation of this optimization problem  (see  also \cite{Sahbi2013icvs, Sahbi2011, Sahbi2014,sahbi2010context,yuan2012mid}), considers handcrafted context in kernel design which makes it possible to clearly enhance the performance of image annotation. As an extension of this work,  we  consider instead a  {\it learned context} as a part of our kernel design framework 
using deep learning \cite{Hinton2006, Benjio2007, Bengio2009, Krizhevsky2012}. Indeed, as an alternative to the fixed context, we consider a deep  network (DN) whose weights correspond to the learned context; high weights in this network characterize the most discriminant parts of the learned context that favor particular spatial geometric relationships  while small weights attenuate other relationships. Context update (i.e., DN weights update)  is achieved using an "end-to-end" framework that back-propagates the gradient of a regularized objective function (which seeks to reduce the classification error of the SVMs built on top of the learned context-aware kernels).  

Note that  context learning  has recently attracted some attention for different tasks;  for instance, 3D holistic scene understanding~\cite{ZhangBaiICCV2107}, scene parsing~\cite{HungTsaiICCV2107} and person re-identification~\cite{HungTsaiCVPR2107} etc., demonstrating that context is an important clue to improve performances. Our contribution is  very different from these works as we consider deep context learning as a part of similarity (and kernel feature) design  for  2D image annotation. Our work is rather more related to  Convolutional Kernel Networks (CKN) \cite{Mairal2014} which learn kernel maps for gaussian functions using convolutional networks. However,  our proposed contribution, in this paper, is   conceptually different from CKN: on the one hand,  our solution is meant to learn a more general class of kernels that   capture context using DN. On the other hand,  the particularity of our work w.r.t CKN (and also w.r.t \cite{Jiu2016b}) is to build   deep kernel map networks while also {\it learning and optimizing   context}.

The rest of this paper is organized as follows: first, we briefly remind in Section~\ref{sec:unslearning} our previous context-aware kernel map learning \cite{Sahbi2013icvs, Sahbi2015}  and we introduce our new contribution in Section~\ref{sec:deepconstruction}: a deep network that allows us to design  kernels and enables us to automatically learn better context. The experimental results on the ImageCLEF annotation benchmark are shown in Section~\ref{sec:experiments}, followed by conclusions in Section~\ref{concl}.

\section{Context-aware kernel maps}\label{sec:unslearning}

Let $\{{\cal I}_p\}_p$ be a collection of images and let ${\cal S}_p=\{ {\bf x}_1^p, \ldots, {\bf x}_n^p \}$ be a list of non-overlapping cells taken from a regular grid of ${\cal I}_p$; without a loss of generality, we assume that $n$ is constant for all images. We measure the similarity between any two given images ${\cal I}_p$ and ${\cal I}_q$ using  the convolution kernel, which is defined as the sum of the similarities  between all the pairs of cells  in ${\cal S}_p$ and ${\cal S}_q$:  ${\cal K}( {\cal S}_p, {\cal S}_q) = \sum_{i, j} \kappa ({\bf x}_i^p, {\bf x}_j^q)$. Here $\kappa ({\bf x}_i^p, {\bf x}_j^q)$ is a symmetric  and positive semi-definite (p.s.d) function that  returns the similarity between two cells. Resulting from the closure of the p.s.d with respect to the convolution, the latter is also positive semi-definite.  In this definition, $\kappa$ usually corresponds to standard kernels (such as polynomial or gaussian) which only rely on the  visual content of the cells; in other words, cells 
are compared independently.   As context may also carry out discriminant clues for classification,  a {\it more relevant} kernel $\kappa$ should provide a high similarity between  pairs of cells, not only when they share  similar content, but also similar context. 

\indent  Following our previous work~\cite{Sahbi2013icvs, Sahbi2015}, our goal is to learn  the kernel $\kappa$ (or equivalently its gram matrix  $\mathbf{K}$) that returns similarity between all data in  ${\cal X}=\cup_p {\cal S}_p$. The objective function, used to build ${\mathbf{K}}$,  is defined as 
\begin{equation}
 \min_{\mathbf{K}} \t(-\mathbf{K}\mathbf{S}') - \alpha \sum_{c=1}^C \t (\mathbf{K} \mathbf{P}_c \mathbf{K}' \mathbf{P}_c^{'} ) + \frac{\beta}{2} ||\mathbf{K}||^2_2,
\label{equa:kernelfunction}
\end{equation}

\noindent here $\beta > 0$, $\alpha \geq 0$, $\kappa({\mathbf{x}, \mathbf{x}'})=\mathbf{K}_{\mathbf{x}, \mathbf{x}'}$ (with $\mathbf{K}_{\mathbf{x}, \mathbf{x}'}$ being an entry of $\mathbf{K}$), $\mathbf{S}$ is a (context-free) similarity matrix between data in $\cal X$, $'$ and $\t$  denote matrix transpose and the trace operator respectively. The first term of Eq. (\ref{equa:kernelfunction}) seeks to maximize kernel values between any pair of cells  ${\mathbf{x}, \mathbf{x}'}$ with a high similarity $\mathbf{S}_{\mathbf{x}, \mathbf{x}'}$ while the second term aims to maximize kernel values between cells whose neighbors are highly similar too (and vice-versa). Finally, the third term acts as a regularizer that also helps getting a smooth closed-form kernel solution (see details subsequently). 
 
\begin{figure}[tbp]
\centering
\includegraphics[width=0.3\linewidth]{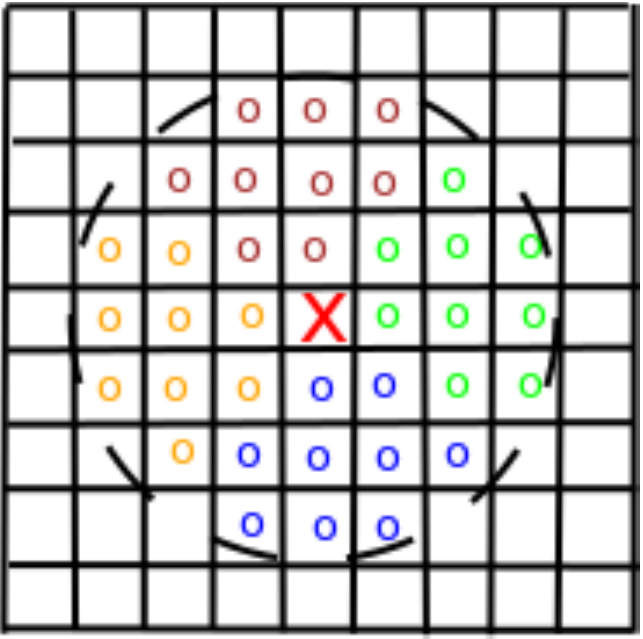}
\caption{\small This figure shows the handcrafted neighborhood system used in order to define the context-aware kernels. Red cross stands for a particular cell  in the regular grid, and colored circles stand for its 4 different types of neighbors. Here we consider a disk with 4 sectors.}
\label{fig:imagegrids}
\end{figure}

\noindent In the above objective function,  the context of a cell refers to the set of its neighbors (see also Fig.~\ref{fig:imagegrids}); the intrinsic adjacency matrices $\{\mathbf{P}_c\}_c$  model  this context between cells. More precisely,  for a given $c \in \{ 1, \dots,  C\}$ ($C=4$ in practice),  the matrix $\mathbf{P}_c$ captures a particular geometric relationship between neighboring cells; for instance when $c=1$,   $\mathbf{P}_{c,\mathbf{x}, \mathbf{x}'}\leftarrow 1$ iff $\mathbf{x}$, $\mathbf{x}'$ belong to the same image and $\mathbf{x}'$ corresponds to one of the {\it left} neighbors of  $\mathbf{x}$  in the regular grid, otherwise  $\mathbf{P}_{c,\mathbf{x}, \mathbf{x}'} \leftarrow 0$.

\indent  We can show that the optimization problem is Eq.~(\ref{equa:kernelfunction}) admits the following closed-form kernel solution (details about the proof are omitted  in this paper) 
\begin{equation}
\mathbf{K}^{(t+1)} = \mathbf{S} + \gamma \sum_{c=1}^C \mathbf{P}_c \mathbf{K}^{(t)} \mathbf{P}_c^{'},
\label{equa:kernelsolution} 
\end{equation}

\noindent here $\mathbf{K}^{(0)}=\mathbf{S}$, $\gamma = \alpha / \beta$ with $\alpha / \beta$ chosen in order to guarantee the convergence of the closed-form solution (\ref{equa:kernelsolution})  to a fixed-point (which is always observed when $\gamma$ is not very large\footnote{i.e., $\gamma$ is chosen to make the norm of the right-hand side term in Eq. \ref{equa:kernelsolution} not very large compared to the left-hand side term.}).  Note that when $\mathbf{S}$ is p.s.d, all the resulting closed-form solutions (for different $t$)  will also be p.s.d and this results from the closure of the positive semi-definiteness w.r.t. different operations including sum and product. Hence, this kernel solution can be expressed as an inner product $\mathbf{K}^{(t+1)} =\mathbf{\Phi}^{(t+1)'}  \mathbf{\Phi}^{(t+1)}$ involving maps that take data from their input space to a high dimensional space; one may show that this map $\mathbf{\Phi}^{(t+1)}$  is explicitly and recursively given as    

\begin{equation}
\mathbf{\Phi}^{(t+1)} = \Big( \mathbf{\Phi}^{'(0)} \  \   \gamma^{\frac{1}{2}} \mathbf{P}_1 \mathbf{\Phi}^{'(t)} \ \ldots \  \  \gamma^{\frac{1}{2}} \mathbf{P}_C \mathbf{\Phi}^{'(t)} \Big)'.
\label{equa:mapsolution}
\end{equation}
\noindent Here ${\bf \Phi}^{(0)}_\x= {\bf V}_\x$ (with ${\bf S}_{\x,\x'}={\bf V}'_\x {\bf V}_\x$) and the subscript in ${\bf {\Phi}}_{\x}$ denotes the restriction of the map ${\bf {\Phi}}$ to a point $\x$. According to Eq.~(\ref{equa:mapsolution}), it is clear that the mapping ${\bf \Phi}^{(t+1)}$ is not equal to  ${\bf \Phi}^{(t)}$ since the dimensionality of the map increases w.r.t. $t$. However, the convergence of the inner product  ${\bf \Phi'}^{(t+1)}{\bf \Phi}^{(t+1)}$ to a fixed-point is again guaranteed when $\gamma$ is bounded, i.e., the gram matrices of the designed kernel maps are convergent. \\  
\noindent Resulting from the definition of the adjacency matrices  $\{{\bf P}_c\}_c$ and from Eq.~(\ref{equa:mapsolution}), it is easy to see that building kernel maps could be achieved image per image with obviously the same number of iterations, i.e., the evaluation of kernel maps of a given image is independent from the others and hence not transductive.

\begin{figure}[htbp]
\centering
\scalebox{0.45}{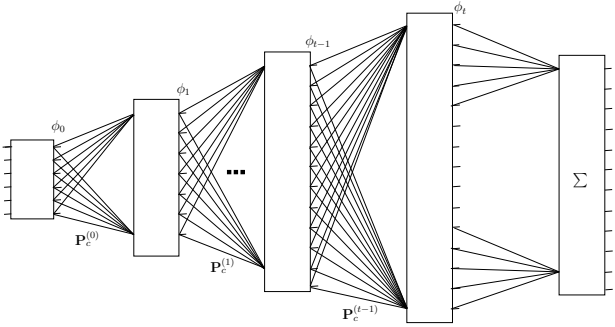}
\caption{\small This figure shows that the learned kernel maps can be seen as multi-layered maps of increasing dimensionality  corresponding to larger and more influencing geometrical context.}\label{deep1}
\end{figure}

\begin{figure*}[htbp]
\begin{center}
\includegraphics[scale=0.35]{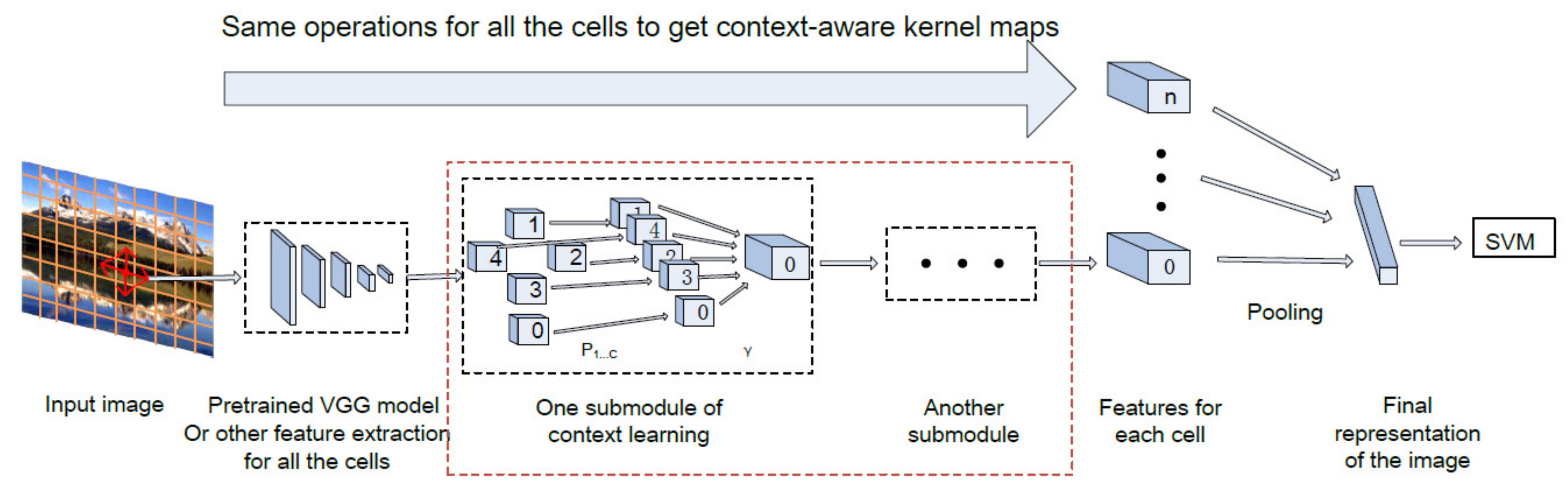}
\end{center}
\caption{\small \textcolor{black}{This figure shows the whole architecture and flowchart of our deep context learning.  Given an input image (divided into $8\times10$ cells; as also shown in  experiments),  cells are first described using the pre-trained VGG CNN. Afterwards, the context-based kernel map  of a given cell (for instance cell 0), at a given iteration, is obtained by combining the kernel maps of its neighboring cells (namely cells 1, 2, 3 and 4), obtained at the previous iteration, as shown in the red dashed rectangle and also in Eq.~(\ref{equa:mapsolution}). At the end of this iterative process,  the kernel maps of all the cells are pooled together in order to obtain the global representation of the input image, prior to achieve its classification. Note that  the network  shown in the red rectangle, together with the pooling layer, correspond   to the DN shown in Fig. \ref{deep1}.}} \label{fig:flowchart}
\end{figure*}

\section{Context Learning with Deep Networks} \label{sec:deepconstruction}

The framework presented in Section~\ref{sec:unslearning} is able to design more effective kernels (compared to context-free ones)  by taking into account  the context. However,  this framework is totally unsupervised, so it does not benefit from existing labeled data in order to produce more discriminating kernels; furthermore, the context (and its matrices $\{\mathbf{P}_c\}_c$) is completely handcrafted. Considering these issues, we propose in this section a method that considers context learning as a part of kernel design using deep networks.  

\subsection{From context-aware kernels to deep networks} 

Considering any two samples $\x$, $\x'$ in $\cal X$ and also the kernel definitions in Eqs.~(\ref{equa:kernelsolution}) and (\ref{equa:mapsolution}), one may rewrite  ${\K}_{\x,\x'}^{(t)}$ as 
\begin{equation*} 
\begin{array}{l}
{\K}_{\x,\x'}^{(t)}=\underbrace{\phi_t(\phi_{t-1}(...\phi_1(\phi_0(\x))))}_{t \ \textrm{times}} . \underbrace{\phi_t(\phi_{t-1}(...\phi_1(\phi_0(\x'))))}_{t \ \textrm{times}}, 
\end{array}
\end{equation*}

\noindent with $\phi_{t}(\x)=\PPhi_\x^{(t)}$. Following  the definition of the convolution kernel ${\cal  K}$, at iteration $t$, we have 

\begin{equation}
\begin{array}{ll}\label{sck}
 \displaystyle {\cal  K}( \S_p, \S_q)  & =  \displaystyle  \sum_{\x \in \S_p} \sum_{\x' \in \S_q} {\K}_{\x,\x'}^{(t)}    \\ 
                           & = \displaystyle \sum_{\x \in \S_p}  \underbrace{\phi_t(\phi_{t-1}(...\phi_1(\phi_0(\x))))}_{t \ \textrm{times}}  \\ 
                    &  \displaystyle \ \  \ \ \  \ \ \ \ \ . \sum_{\x' \in \S_q}  \underbrace{\phi_t(\phi_{t-1}(...\phi_1(\phi_0(\x'))))}_{t \ \textrm{times}}.
\end{array}
\end{equation} 
 
Each side of the convolution kernel, in Eq. (\ref{sck}), corresponds to multi-layered feature maps of increasing dimensionality that capture larger and more influencing context, related to points and their geometrical configurations (see \textcolor{black}{Fig.~\ref{deep1}}). We observe that the architecture in this figure is very similar to the one widely used in deep learning, with some differences residing in the definition of the weights. Indeed, the latter correspond to the values of matrices $\{\P^{(t)}_c\}_{t,c}$ for different iterations and the weights in the final layer correspond to pooling (summation) weights; the latter are equal to $1$ following Eq.~(\ref{sck}). In contrast to many usual deep learning algorithms, the weights of  this {deep network} are easy to interpret.

\noindent Note that the {\it appropriate}  number of layers in this network is defined by the asymptotic behavior of our context-aware kernel; indeed, the number of iterations necessary in order to obtain convergence is exactly the number of layers in this deep network. In practice, convergence happens in less than five iterations \cite{Sahbi2015}. Now, considering ${\bf \tilde{K}}$ as the limit of Eq.~(\ref{equa:kernelsolution}) (for some $t \leadsto T$) and ${\bf \tilde{\Phi}}$ as the underlying kernel map (using Eq.~(\ref{equa:mapsolution})), the new form of the convolution kernel $\cal K$  between two sets of points ${\cal S}_p$, ${\cal S}_q$ can be rewritten  ${\cal K}({\cal S}_p,{\cal S}_q) =  \sum_{(\x,\x') \in {\cal S}_p \times {\cal S}_q } \langle {\bf \tilde{\Phi}}_{\x},{\bf \tilde{\Phi}}_{\x'} \rangle$. It is easy to see that $\cal K$ is a p.s.d kernel as it can be rewritten as a dot product involving finite dimensional and explicit maps, i.e.,  ${\cal 
K}({\cal S}_p,{\cal S}_q) =  \langle \phi_{\cal K}({\cal S}_p), \phi_{\cal K}({\cal S}_q) \rangle$, with $\phi_{\cal K}({\cal S}_p) = \sum_{\x \in {\cal S}_p}  {\bf \tilde{\Phi}}_{\x}$, which clearly shows that each constellation of  points ${\cal S}_p$ can be {\it deeply} represented with the explicit kernel map $\phi_{\cal K}({\cal S}_p)$ (i.e., the output of the deep network in the \textcolor{black}{Fig.~\ref{deep1}}). 

It is worth noticing that our context learning framework, in spite of being targeted to kernel design,  can  be achieved  "end-to-end" thanks to the existence of the maps $\phi_0(.)$ for many kernels (e.g. linear, histogram intersection, polynomial kernel and other complex kernels); see for instance \cite{Sahbi2013icvs}. Therefore,   we adopt  back-propagation in order to update the weights of the network in Fig.~\ref{deep1} as described subsequently (see also the flowchart of the whole framework in Fig.~\ref{fig:flowchart}); interestingly, this framework could also benefit from pre-trained convolutional neural networks (CNN) as input to the kernel map $\phi_0(.)$  as also shown   through experiments in  Section~\ref{sec:experiments}. 

\begin{table*}[t]
	\centering
\resizebox{0.8\textwidth}{!}{	
	\begin{tabular}{|c|c|cc|cc|}
	\hline
	\multirow{2}{*}{$r$} & \multirow{2}{*}{Method} & \multicolumn{2}{c|}{BoW features} & \multicolumn{2}{c|}{VGG-CNN features} \\
        \cline{3-6}
	& & Linear kernel  & HI kernel & Linear kernel&  HI kernel \\
	\hline
	& Context-free & 39.7/24.4/46.6 & 41.3/25.1/49.5 & 45.3/30.8/56.4 & 45.5/30.1/57.9  \\
	\hline
	\multirow{2}{*}{1}  & fixed Context-aware  (\cite{Sahbi2013icvs}) & 40.6/24.6/48.3 & 42.6/26.3/50.5 & 45.8/31.2/57.6 & 46.4/30.7/58.5  \\
	& learned Context-aware  & 42.7/26.4/50.5 & 45.2/26.4/53.9 & 47.5/32.7/58.7 & \textbf{48.8/32.7/59.9} \\
	\hline
	\multirow{2}{*}{5} & fixed Context-aware (\cite{Sahbi2013icvs}) & 41.0/25.3/48.9 & 42.9/26.7/51.3 & 46.8/31.8/57.9 & 46.9/31.1/58.7  \\
	& learned Context-aware & \textbf{44.0/26.6/52.0} & \textbf{45.6/26.2/54.0} & \textbf{47.9/33.2/58.8} & 48.4/32.7/59.5 \\
	\hline
	\end{tabular} 
	}
	\vspace{0.1cm}
	\caption{\small \textcolor{black}{The performance (in $\%$) of different kernels on ImageCLEF database. A triple $\cdot/\cdot/\cdot$ stands for MFS/MFC/MAP. In these experiments $r$ corresponds to the radius of the disk that supports context.} \label{tab:imageclefresults}}
\end{table*} 

\subsection{Context learning} 
Considering a multi-class problem with $K$ classes and $N$  training images $\{{\cal I}_p\}_{p=1}^N$, we define  $\Y_k^p$  as the membership of the image  ${\cal I}_p$ to the class $k  \in \{1,\dots,K\}$; here  $\Y_k^p=+1$ iff ${\cal I}_p$  belongs to class $k$ and  $\Y_k^p=-1$ otherwise.  We consider in this section, a dynamic and discriminative update of  matrices  $\{\P_c^{(t)}\}_{c,t}$; first, we  plug the explicit form of $\tilde{\PPhi}$ into $\phi_\KK({\cal S}_p)$, then we optimize the following objective function (w.r.t.  $\{\P_c^{(t)}\}_{c,t}$ with $t=0,\dots,T-1$ and the SVM parameters denoted $\{w_k\}_k$)\footnote{For ease of writing and unless confusing, the superscript $t$ is sometimes omitted in the notation.}. This objective function includes the following  regularization term and hinge loss
\begin{equation}\label{eq8}
\min_{\P_c,w_k}   \displaystyle   \sum_{k=1}^K \frac{1}{2} ||w_k||^2 + C_{k}  \sum_{p=1}^{N} \max(0, 1-\Y_k^p w_{k}' \phi_\KK({\cal S}_p)),  
\end{equation} 

\noindent as the optimization of  the above objective function w.r.t. the two sets of parameters together is difficult, we adopt an alternating optimization procedure; this is iteratively achieved by fixing one of the two sets of variables and solving w.r.t. the other. When fixing $\{\P_c^{(t)}\}_{c,t}$, the goal is to learn the parameters  $\{w_k\}_k$ of $K$   binary SVM classifiers (denoted $\{f_k\}_{k=1}^K$). Since kernel map $\phi_\KK({\cal S}_p)$  of a given image ${\cal I}_p$ is explicitly given as the output of the deep network in Fig.~\ref{deep1}, each classifier $f_k$ can be written as $f_k({\cal I}) = w_k' \phi_\KK({\cal S}_p)$, where $w_k$ corresponds to the parameters of the SVM. The optimization of Eq.~(\ref{eq8})  w.r.t. these  parameters $\{w_k\}_k$  is achieved using LIBSVM \cite{Chang2011}. 

When fixing $\{w_k\}_k$, the optimization of Eq.~(\ref{eq8}) is  achieved (w.r.t. $\{\P_c^{(t)}\}_{c,t}$) using gradient descent. Let ${E}$ denote the objective function in Eq.~(\ref{eq8}), the gradient of ${E}$ w.r.t. the final kernel map $\phi_\KK$ (i.e., output of the DN) is 
\begin{equation}
	\frac{\partial E}{\partial \phi_\KK} = - \sum_{p=1}^N \sum_{k=1}^K C_k \Y_k^p w_k \mathbbm{1}_{\{1-\Y_k^p w_{k}' \phi_\KK({\cal S}_p)\}},
\label{equa:gradientl1}
\end{equation}

\noindent here $\mathbbm{1}_{\{\}}$ is the indicator function. From Eq.~(\ref{equa:mapsolution}), it can be seen that the gradient of ${E}$ w.r.t. $\{\P_c^{(T-1)}\}_{c,T-1}$ can be computed as the sum of gradients over all the cells in the regular grid. This gradient is backward propagated to the previous layers using the chain rule \cite{LeCun98} in order to (i) obtain  gradients w.r.t.   adjacency matrices $\{\P_c^{(t)}\}_{c,t}$, for $t=T-1,\dots, 0$  (weights of the deep network in Fig.~\ref{deep1}), and (ii) update these weights   using gradient descent. \\
\indent Finally, the two steps of this iterative process are repeated till convergence is reached, i.e., when the values of these two sets of parameters remain unchanged. In practice, this convergence is observed in less than 100 iterations. 

\section{Experimental Validation} \label{sec:experiments}

We evaluate the performance of our context and kernel learning method on the challenging ImageCLEF benchmark~\cite{Villegas2013}. The targeted task is image annotation; given an image, the goal is to predict the list of concepts present into that image. These concepts are declared as present iff the underlying SVM scores are positive. 

The dataset of this benchmark is large and includes more than 250k images; we only use the dev set (which includes 1,000 images) as the ground truth is released for this subset only. Images of this subset belong to 95 categories (concepts); as these concepts are not exclusive, each image may be annotated with one or multiple concepts. We randomly split the dev set into two subsets (for training and testing).  We process each image in the training and testing folds by re-scaling its dimensions to $400 \times 500$ pixels\footnote{This corresponds to  the median dimension of images in the dev set.} and partitioning its spatial extent into $8 \times 10$ cells; each cell includes  $50 \times 50$ pixels. \textcolor{black}{Two different representations are adopted in order to describe each cell: i) Bag-of-Words (BoW) histogram (of 500 dimensions) over SIFT features, where the codewords of this histogram are pre-trained  offline using \textit{K}-means; and ii) Deep features  based on  the pre-trained VGG model on the ImageNet database  (``imagenet-vgg-m-1024'') \cite{Chatfield14}. Our purpose here is to investigate the adaptability of the proposed framework both with handcrafted and pre-trained  CNN features.}

\begin{table}[htb]
	\centering
\resizebox{0.45\textwidth}{!}{	
	\begin{tabular}{c|ccc}
	\hline
	Kernel & MF-S & MF-C & mAP \\
	\hline
    \hline
    GMKL(\cite{Varma2009})  & 41.3 & 24.3 & 49.1 \\
    2LMKL(\cite{Zhuang2011a}) & 45.0 & 25.8 & 54.0 \\
    LDMKL (\cite{Jiutip2017}) & 47.8 & 30.0 & 58.6  \\
    The proposed context-aware & 48.8 & 32.7 & 59.9  \\ 	
	\hline
	\end{tabular}
	}
    \vspace{0.1cm}
    \caption{\small \textcolor{black}{This table shows comparison of performances (in \%) between different kernel-learning  methods.} \label{tab:comparsionresults}}
\end{table}

\begin{figure*}[thbp]
\begin{center}
\includegraphics[angle=0,width=1\linewidth]{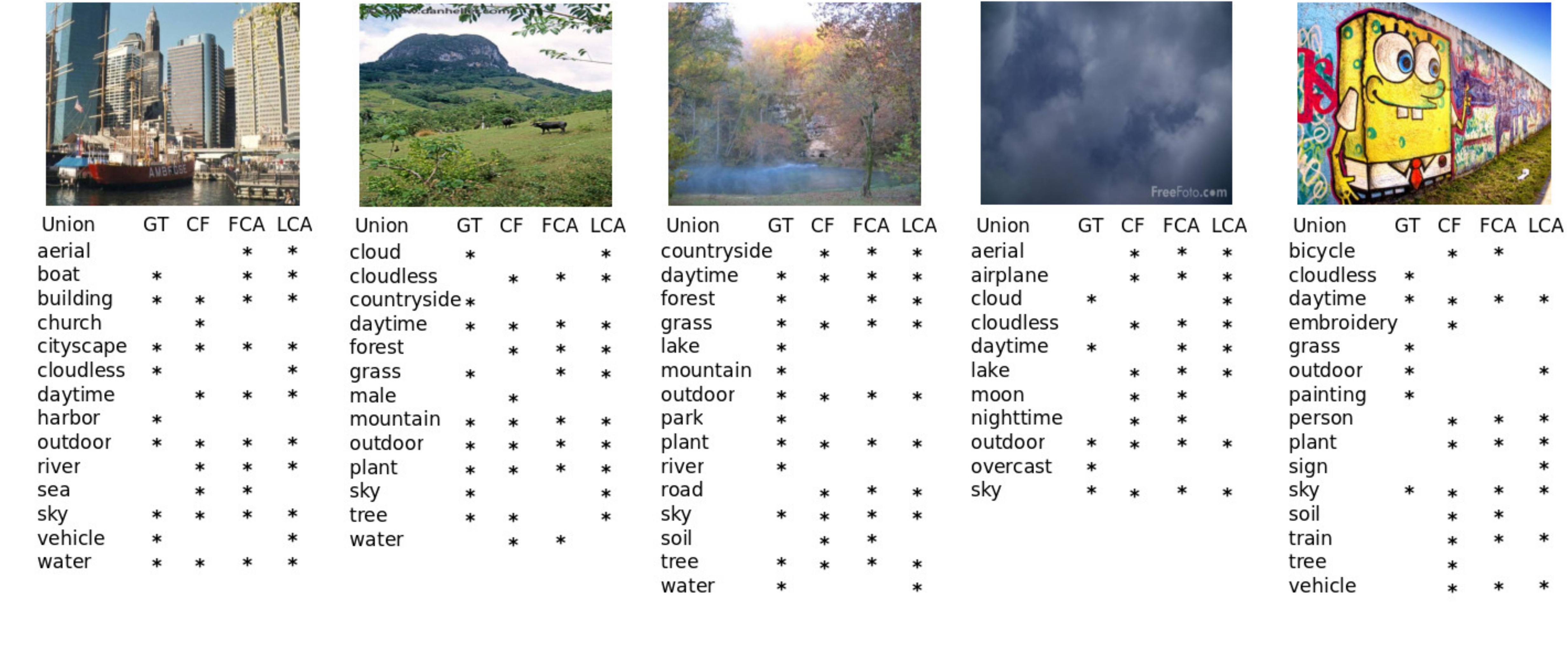}
\end{center}
\caption{\small Examples of annotation results using context-free kernels (``CF''), context-aware kernels with fixed and learned context (resp. ``FCA'' and "LCA"). ``GT'' refers to ground truth annotation while  the stars stand for the presence of a given concept in the test image.} \label{fig:annotationexamples}
\end{figure*}

\begin{figure}[thbp]
\begin{center}
\includegraphics[scale=0.33]{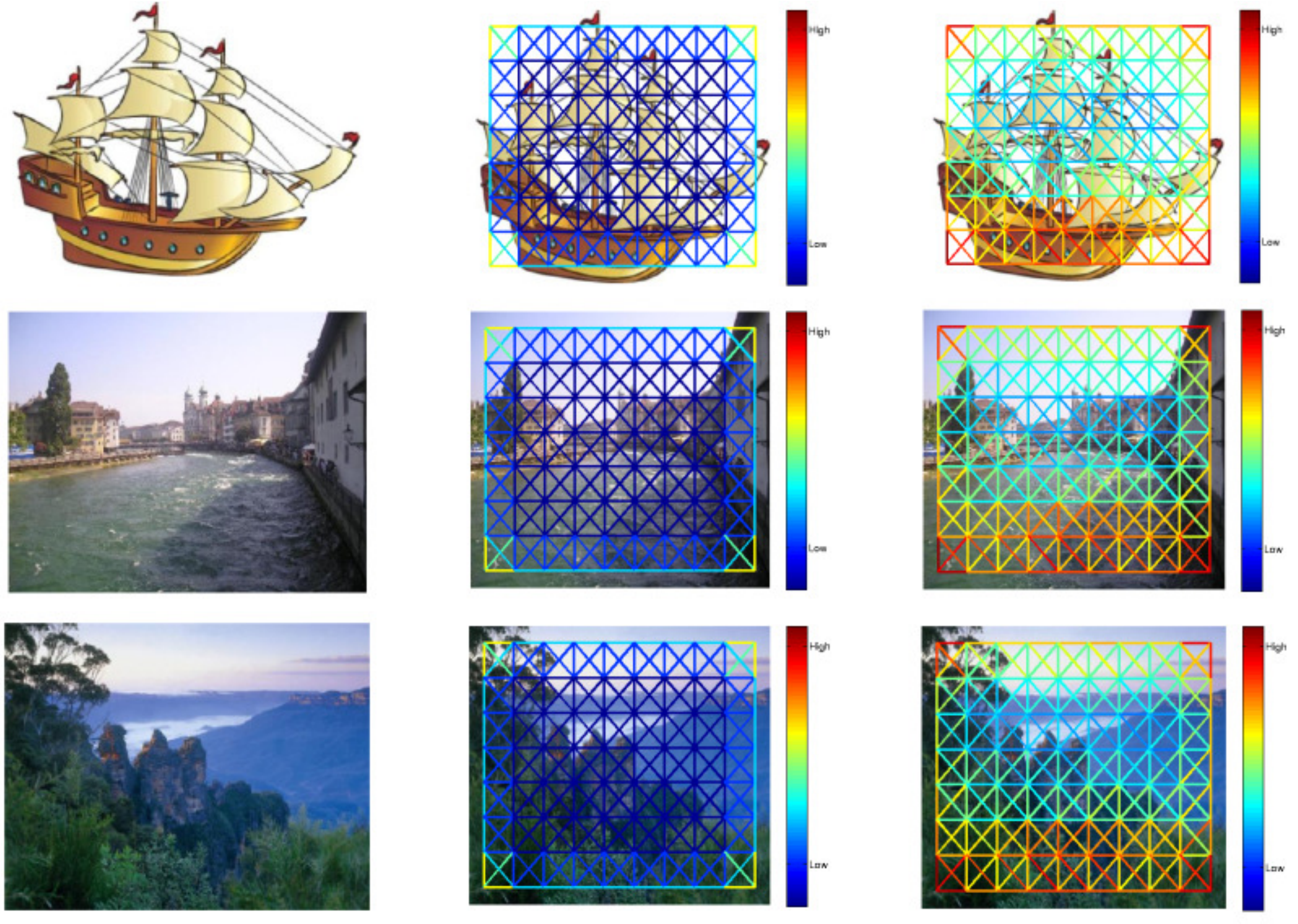}
\end{center}
\caption{\small This figure shows original images (left), handcrafted (middle) and learned contexts (right) between $8\times10$ cells (with $r=1$). For each cell, the importance of its context is shown with colored connections to its neighboring cells; warm colors stand for important  relationships  while cold colors stand for less important ones. In the handcrafted context, adjacency matrices  $\{\mathbf{P}_{c}\}_c$, are set to be row-stochastic, and this is obtained by normalizing each row (cell) in these matrices by the number of  its spatial neighbors  that's why cells in the four corners have larger values (better viewed/zoomed in PDF)} \label{fig:contextexamples}
\end{figure}

We consider four types of geometric relationships in order to build our context-aware kernel maps (see again Fig. \ref{fig:imagegrids}). As the dimensionality of the DN (in Fig.  \ref{deep1}) increases rapidly with the layers, we adopt in our experiments an  architecture depth that provides a reasonable balance between dimensionality and performance as well as convergence of kernel map evaluation\footnote{Again as mentioned earlier in Section~\ref{sec:unslearning},  $\gamma$ and the number of iterations $T$ should be set appropriately in order to obtain convergence of kernel evaluation; in practice, with $\gamma=1$ convergence is well approached with only $T=3$ iterations of kernel map evaluation ($T+1$ is then the chosen number of layers in our DN).}; hence, we chose an architecture with 3+1 layers where the last one corresponds to pooling  (again following Fig.~\ref{deep1}). In these experiments, we  consider linear and histogram intersection maps as -- context-free kernel map --  initialization of our DN;  note that {the explicit maps of linear and histogram intersection kernels correspond to identity and decimal-to-unary mapping respectively (see~\cite{Sahbi2015} for more details about these maps).} 

\indent Using the setting above, we learn multi-class SVMs for different concepts (using the training fold) on top of the learned context-aware kernel maps, and we evaluate their performances on the testing fold. These performances are measured using the F-scores (harmonic means of recall and precision) at the sample and concept levels (denoted MFS and MFC respectively) as well as the mean average precision (MAP); higher values of these measures imply better performances. 
Tab.~\ref{tab:imageclefresults} shows a comparison of our context-aware kernels against context-free ones, with handcrafted and learned contexts on top of the BoW features \textcolor{black}{as well as the deep VGG features}. From these results the gain obtained with the learned context is clear both w.r.t. handcrafted context and when no context is used. \textcolor{black}{Tab.~\ref{tab:comparsionresults} shows   performance comparison  of our learned context-aware kernel  against other kernel learning methods on the ImageCLEF database; from these results a clear gain is obtained when learning context.}  Fig.~\ref{fig:annotationexamples}  shows a sample of images and their annotation results  using context-free and context-aware kernels with fixed (handcrafted) and learned context. Finally, Fig.~\ref{fig:contextexamples} is a visualization of handcrafted and learned contexts superimposed on three images from ImageCLEF; when context is learned, it is clear that some spatial cell context relationships are {\it amplified} while others are {\it attenuated} and this reflects their importance in kernel learning and image classification.  

\section{Conclusion}\label{concl}
We introduced in this paper a novel method that considers context learning as a part of kernel design. The method is based on a particular deep network architecture which corresponds to the map of our context-aware kernel solution. In this contribution, relevant context is selected by learning weights of a DN; indeed high weights correspond to the relevant parts of the context while small weights are related to the irrelevant parts. Experiments conducted on the challenging ImageCLEF benchmark, show a clear gain of SVMs trained on top of kernel maps (with learned context) w.r.t. SVMs trained on top of kernel maps (with handcrafted context) and context-free kernels. 

\section*{ACKNOWLEDGMENT}

{This work was partially supported by a grant from the research agency ANR (Agence Nationale de la Recherche) under the MLVIS project (Machine Learning for Visual Annotation in Social-media, ANR-11-BS02-0017). Mingyuan Jiu was with T\'el\'ecom ParisTech as a postdoc, working with Hichem Sahbi, when this work was discussed and part of it achieved and written (under the MLVIS project).}

\bibliography{paper}

\end{document}